%% file: main.tex
\documentclass{INTERSPEECH2023}

\newcommand \ignore[1]{}

\usepackage{url}
\usepackage{hyperref}
\usepackage[table,x11names]{xcolor}
\usepackage{caption}
\usepackage{subcaption}
\usepackage{comment}
\definecolor{mycolor}{rgb}{1,0,1} 

\interspeechcameraready

\title{New Disease Detection with Contrastive Learning}
\title{Discovery of COVID-19 Coughing and Breathing Patterns from Flu and Healthy People's Coughing Patterns Using Contrastive Learning}
\title{Discovering COVID-19 Coughing and Breathing Patterns from Non-COVID Coughs Using Contrastive Learning with Varying Pre-Training Domains and Augmentations}
\title{Discovering COVID-19 Coughing and Breathing Patterns from Non-COVID Coughs Using Contrastive Learning with Varying Pre-Training Domains}
\title{Discovering COVID-19 Coughing and Breathing Patterns from Unlabeled COVID-19 Data and Labeled Non-COVID Coughs Using Contrastive Learning with Varying Domains for Pre-Training\ignore{\\ \textcolor{red}{Use BLUE/Other Colors for Anything you modify/add}}}
\title{Discovering COVID-19 Patterns from Unlabeled Data and Labeled Non-COVID Coughs Using Contrastive Learning with Varying Domains for Pre-Training}
\title{Discovering COVID-19 from Unlabeled COVID-19 and Labeled Non-COVID Data Using Contrastive Learning with Varying Pre-Training Domains}
\title{Discovering COVID-19 Coughing and Breathing Patterns from Non-COVID Coughs Using Contrastive Learning with Varying Domains for Pre-Training}
\title{Discovering COVID-19 Coughing and Breathing Patterns from Unlabeled Data Using Contrastive Learning with Varying Pre-Training Domains}

\name{Jinjin Cai$^1$\thanks{This activity was funded by Purdue University in support of the West Lafayette-Indianapolis initiative aimed at strengthening collaboration between Purdue units on the two campuses.}, Sudip Vhaduri$^1$, Xiao Luo$^2$}
\address{
  $^1$Comp \& Info Tech Dept, Purdue University, West Lafayette, IN 47906, USA\\
  $^2$Comp \& Info Tech Dept, IUPUI, Indianapolis, IN 46202, USA}
\email{$^1$\{cai379,svhaduri\}@purdue.edu, $^2$luo25@iupui.edu}

\begin{document}

\maketitle
 
\begin{abstract}
Rapid discovery of new diseases, such as COVID-19 can enable a timely epidemic response, preventing the large-scale spread and protecting public health. However, limited research efforts have been taken on this problem. In this paper, we propose a contrastive learning-based modeling approach for COVID-19 coughing and breathing pattern discovery from non-COVID coughs. To validate our models, extensive experiments have been conducted using four large audio datasets and one image dataset. We further explore the effects of different factors, such as domain relevance and augmentation order on the pre-trained models. Our results show that the proposed model can effectively distinguish COVID-19 coughing and breathing from unlabeled data and labeled non-COVID coughs with an accuracy of up to $0.81$ and $0.86$, respectively. Findings from this work will guide future research to detect an outbreak of a new disease early. 

\end{abstract}

\input{introduction}

\input{approach}

\input{analysis}

\input{discussion}

\bibliographystyle{IEEEtran}
\bibliography{reference}

\end{document}

%% file: introduction.tex
\section{Introduction}\label{introduction}

\subsection{Motivation}\label{motivation}
Rapid discovery of new diseases is crucial for preventing the spread of infectious diseases and protecting public health \cite{morse2012public}. The past three years have seen the devastating impact of the global COVID-19 pandemic, with over 757 million infections and 7.6 million deaths \cite{who}. The extensive lockdowns and travel restrictions disrupted the global economy, causing massive job losses and driving an estimated 100 million people into extreme poverty \cite{nicola2020socio}. Moreover, some patients still suffer from chronic symptoms like brain fog, chest pain, and breathlessness due to irreversible damage inflicted by the virus \cite{sequece}.
While rapid virus spread presents a significant challenge, early intervention with effective treatments and isolation can drastically stop viral transmission \cite{peck2020early} and minimize the impact on patient bodies \cite{kim2020therapy}. However, current detection methods, such as epidemiological surveillance \cite{ibrahim2020epidemiologic}, laboratory testing \cite{lai2021laboratory}, and clinical diagnosis \cite{clinical}, are resource-intensive and financially burdensome. The outbreak also creates extraordinary stress on healthcare providers, i.e., physicians and nurses \cite{hospitalburden}. Additionally, patients who are required to conduct laboratory testing at specific facilities are facing increasing infection risk, especially in under-resourced areas with limited testing supplies \cite{least}.

\textcolor{black}{Therefore, it is crucial to develop a system that can detect new diseases, such as COVID-19 using symptoms captured by common sensors. This will provide a less invasive and more efficient means of disease patterns discovery, which will enable patients to receive timely and appropriate medical care while reducing the spread of the virus,  as well as reducing the burden on healthcare professionals.}

\subsection{Related Work}\label{realted work}


Respiratory diseases, such as  Chronic Obstructive Pulmonary Disease (COPD) \cite{COPD1} and Asthma \cite{ASTHMA1}, affecting the trachea, bronchus, lungs, and chest are often characterized by different symptoms, including coughing, shortness of breath, and wheezing \cite{ijaz2022towards}. Recently researchers have been using audio recordings to develop AI-based approaches to detect various respiratory diseases \cite{dibbo2021effect,vhaduri2023environment}. 
Nonetheless, given the high transmissibility and numerous variants of the SARS-CoV-2 virus causing COVID-19, the urgency for a reliable early-stage detection model to curb its spread has gained global attention. While most studies have focused on traditional supervised learning models to develop approaches \cite{covid-cough}, these models rely on labeled data. But data labeling requires substantial work and specialized knowledge with training before labeling new data.

\ignore{OLD VERSION: To cope with the widespread prevalence of the virus, the detection and diagnosis of patterns of COVID-19 have been successfully carried out by researchers with traditional machine or deep learning methods. A great many researches have been conducted on binary classification machine learning or deep learning models based on chest X-rays or CT scans . Islam \textit{et. al} also proposed a COVID-19 diagnosis algorithm based on Deep Convolutional Network from cough sounds \cite{covid-cough}. By extracting features from the time-domain, frequency-domain, and mixed-domain of recordings, the detection task was able to achieve the highest accuracy of 97.5\%. However, traditional supervised learning relies on labeled data which requires massive work and expert knowledge. Furthermore, supervised models perform badly in classifying unknown new data which they are not trained with.}


Contrastive learning has become a trend recently because it has the ability to learn from comparing instances by generating positive sets and negative sets in an unsupervised way \cite{le2020contrastive,jaiswal2020survey}. It aims to learn representations in abstract space rather than focusing on tedious details so it has a strong generalization ability and has shown competing performance with supervised learning. SimCLR \cite{chen2020simple} is a widely used contrastive learning model whose performance can achieve astonishing results without any labels in training, and even higher than some of the supervised learning models with much fewer labeled data for linear fine-tuning on large-scaled ImageNet dataset.
Recently, some researchers have made attempts to apply contrastive learning to acoustic event detection \cite{con1,con2,saeed2021contrastive,jiang2020speech,con3}. 
While the models developed by previous studies have shown potential in audio classification tasks, they still required labeled data of all classes, failing to handle unlabeled data. Moreover, despite some advancements in cross-domain contrastive learning within fields such as video action recognition \cite{video}, recommendation system development \cite{recommend1,recommend2}, and image classification \cite{image}, there is a noticeable knowledge gap in audio classification. Specifically, the potential impact of domain relevance on audio classification performance remains largely unexplored.
\subsection{Contribution}\label{contribution}

The main contribution of this work is to present a modeling approach that can help to identify coughing and breathing patterns that are different from patterns associated with known diseases. We consider coughs from healthy people and flu patients as our known classes to develop models that can discover COVID-19 coughing and breathing patterns as new (unknown) patterns. With our detailed analysis of four public audio datasets, we are able to identify unlabeled COVID-19 coughing and breathing patterns with an accuracy of up to $0.81$ and $0.86$, respectively, which shows the promise of our approach. For known labeled classes, we observe an accuracy of up to $0.88$ (Healthy coughs) and $0.89$ (Flu coughs). Thereby, the proposed approach using contrastive learning with audio data recorded from smartphones can be used to identify outbreaks of a new respiratory disease without requiring disease-specific data.
Additionally, we find it is better to develop pre-trained models with data from a domain similar to the test domain and audio augmentations are better than image augmentations. These findings can guide future research to predict the outbreak of a new disease.

%% file: approach.tex
\section{Approach}\label{approach}

\begin{figure}[t]
\centering
\includegraphics[width=3.0in]{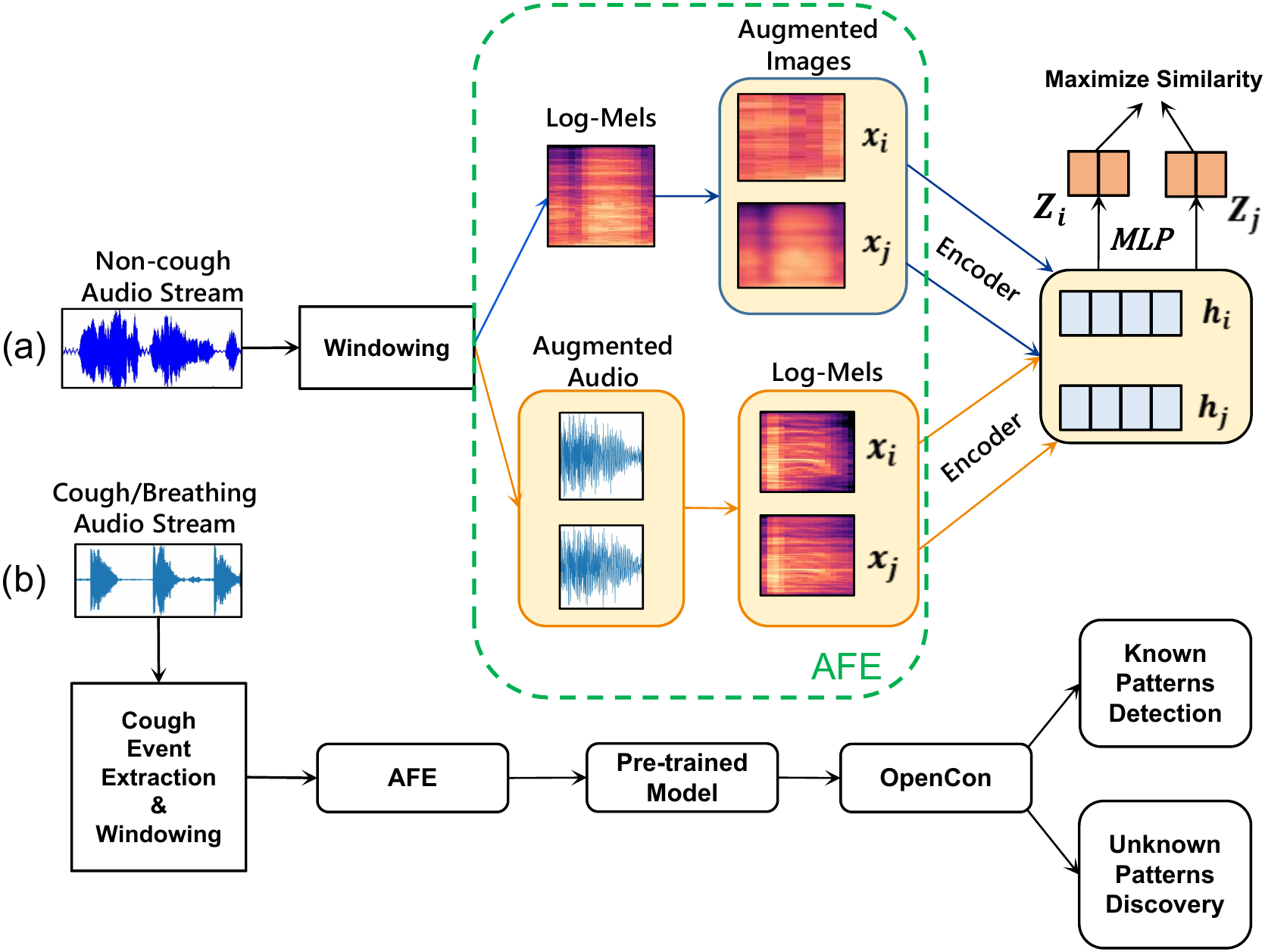}
\caption{Overview of our methods with (a) pre-trained model development and (b) pattern discovery; ``AFE'' stands for augmentation and feature extraction}
\label{model}
\end{figure}

Figure~\ref{model} depicts an overview of our modeling approach, consisting of different processing steps. But before presenting the details, we will first present different datasets used in this work. 

\subsection{Datasets}\label{datasets}
In this work, we use two types of datasets, i.e., image and audio datasets. Throughout this paper, we use the term ``Dataset-n'' to present a dataset and the number of classes (n) used for an experiment/analysis. 


\noindent \textbf{Image Dataset:} The {\em Canadian Institute for Advanced Research} (CIFAR) is a commonly used benchmark dataset for object classification in computer vision research~\cite{cifar10}. 
In this work, we use CIFAR-10 and CIFAR-100, each consisting of 600 images per class, for pre-trained model development. 

\noindent \textbf{Audio Dataset:}
In this work, we use four audio datasets. The {\em Environmental Sound Classification} (ESC) is a widely-used audio dataset with 50 classes of acoustic events, with 40 recordings of 5 seconds per class~\cite{esc50}. 
When developing pre-trained models, we remove the cough and breathing classes and consider the remaining 48 classes (i.e., ESC-48) to avoid any potential bias in the pattern discovery model. 
We used the ESC coughs later when generating representation vectors from the pre-trained models to feed into known healthy cough pattern detection models.  


Our second audio dataset is the {\em AudioSet} dataset, which is a large-scale acoustic dataset that contains 632 kinds of sounds extracted from online videos, with over 2 million human-labeled 10-second audio clips~\cite{audioset}. In this work, we pick a subset of 48 human sound classes from the AudioSet (i.e., AudioSet-48), which are generated from human mouths and noses, such as snoring, sneezing, singing, and crying.
Similarly to ESC, cough and breathing sounds are excluded from pre-trained models and cough sounds are later utilized when generating representation vectors from the pre-trained models to feed into known healthy cough pattern detection models.

{\em Coswara} is our third audio dataset, which is a large COVID-19 dataset of over 20,000 audio recordings from COVID-19-positive and healthy individuals~\cite{coswara} collected by the Indian Institute of Science (IISc). It contains recordings, such as breathing, cough, and speech sounds, from both patients and healthy people. 
When the COVID-19 patient coughing and breathing sounds from this dataset are used for unknown COVID-19 coughing and breathing pattern discovery, healthy people's coughing sounds are used for known healthy cough pattern discovery. 
 
{\em FluSense} is our fourth audio dataset, which is also a large-scale multi-sound dataset recorded in four public waiting areas within the hospital of the University of Massachusetts Amherst to track influenza-related indicators~\cite{flusense}. It contains instances from flu patients and healthy people with cough, sneezing, sniffling, and other sounds.
Patient coughing sounds from this dataset are used when generating representation vectors from the pre-trained models to feed into known flu cough pattern detection models.

In our experiment settings, the four classes -- healthy people's cough (Healthy), flu patient's cough (Flu), COVID-19 patient's cough (CC), and COVID-19 patient's breathing (CB). 
When the ``Healthy'' cough class (obtained from the ESC, AudioSet, and Coswara datasets) and ``Flu'' cough class (obtained from the FluSense dataset) are used for known healthy cough and flu cough pattern detection, the ``CC'' and ``CB'' classes (both obtained from the Coswara dataset) are used for unknown pattern discovery.

\subsection{Data Processing}\label{data processing}
This section presents cough event extraction and labeling approaches from continuous audio recordings and feature extraction methods.

\subsubsection{Cough Event Extraction and Labeling}\label{cough extract}
A cough event is defined as a two- or three-phase action with \ignore{inhalation, phonation, and termination}explosion, decay, and voiced phases~\cite{vhaduri2020nocturnal}. 
A sequence of coughing events that occur consecutively with an interval of no more than two seconds is referred to as a cough episode. Each audio recording downloaded from different datasets often contains one or more cough episodes. Therefore, we first extract the events manually using the Audacity toolbox, and then, we adapted an automatic energy-threshold-based cough event extraction strategy developed by others~\cite{vhaduri2020nocturnal}. 

\begin{figure}[t]
\centering
\includegraphics[width=2.5in]{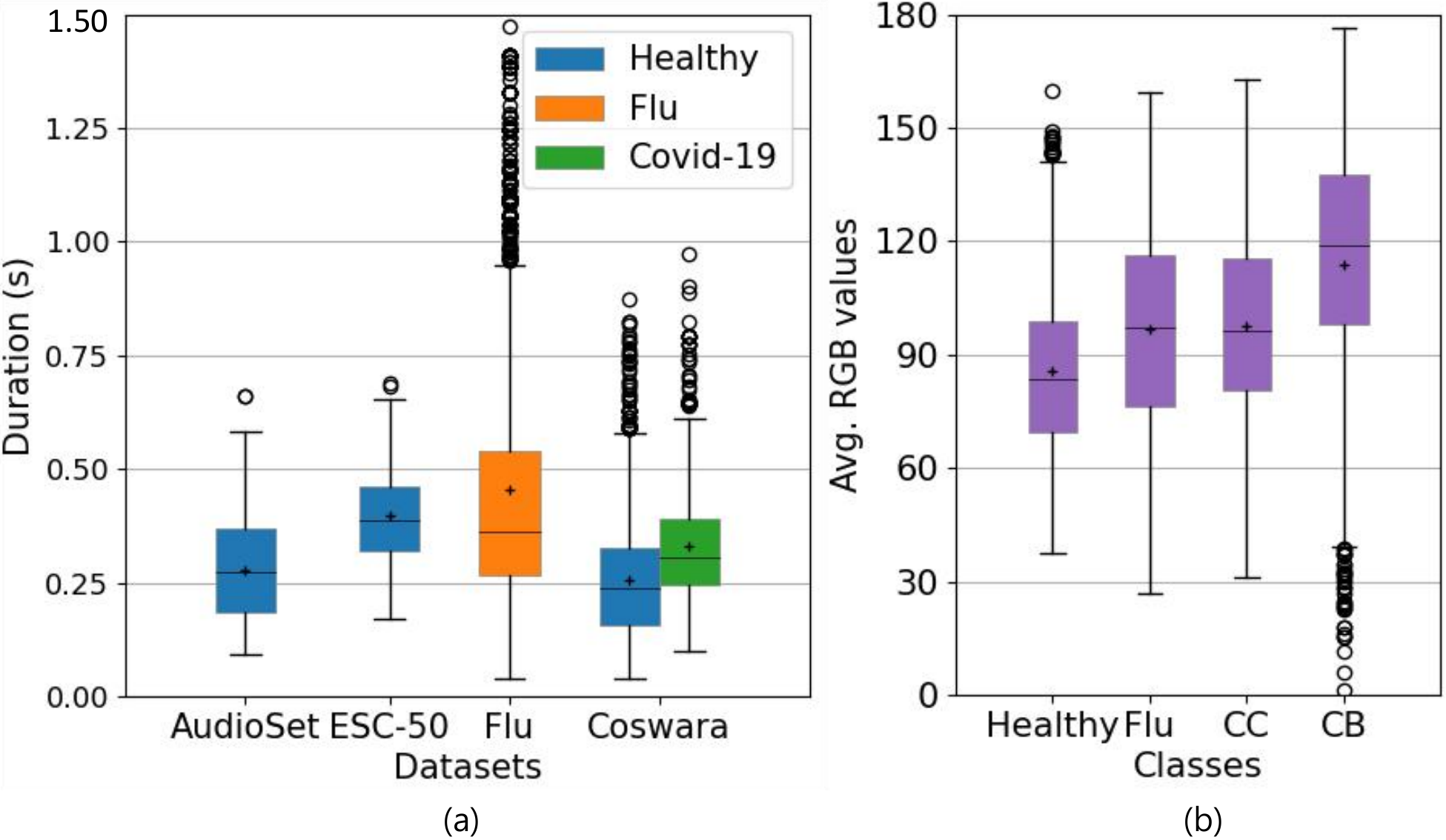}
\caption{Distribution of (a) cough event duration and (b) average RGB values of Log-Mel spectrogram images}
\label{0.5s}
\end{figure}

\subsubsection{Windowing and Feature Extraction}\label{add process}

Since the duration of the cough instances (i.e., events) segmented from different datasets varies, we conduct further processing to standardize the instance or window size. The boxplot in Figure~\ref{0.5s}(a) illustrates the distribution of duration of different types of coughs obtained from various datasets. 
While ``Flu'' coughs are longer and widely varying in lengths, we choose 0.5 seconds as an optimal choice for standard window size. When longer coughs are truncated to 0.5 seconds, shorter coughs are padded with zeros at the end. 
We also consider this 0.5-second window size when segmenting other non-cough audio recordings, including the breathing sounds using non-overlap sliding windows. 


Depending on the order of augmentation (discussed in Section~\ref{methods}), \textit{melspectrogram(.)} function from the \textit{librosa} library is utilized to extract the Log-Mel spectrogram features, a widely used feature for sound classification, from each cough and non-cough window. The \textit{melspectrogram(.)} function performs all signal transformation and filtering to generate the features at a sampling rate of 44100 Hz. 
The extracted spectrograms are then saved as images. Some windows at the beginning or end of an event have less information, i.e., mostly blank or silent, resulting in black areas in the spectrograms. Therefore, to filter out these less informative feature windows, we investigate the distribution of average RGB values of all cough feature images, as presented in Figure~\ref{0.5s}(b). 
We found 70 as our optimal cut-off threshold; with this threshold, we remove 25\% of the cough windows. 
We perform similar processing for the 
non-cough class features used to develop pre-trained models from the ESC-48 and AudioSet-48 datasets.

After all the processing and balancing, we end up with 6000 instances per class from the CIFAR-10 and CIFAR-100 datasets, 380 instances per class from the ESC-48 dataset, and 900 instances per class for AudiSet-48 dataset used for pre-trained model development. 
To generate representation vectors from the pre-trained models, and discover known (``Healthy'' and ``Flu'' cough) patterns and unknown (``CC'' and ``CB'') patterns, we use 1600 instances per class.

\subsection{Methods}\label{methods}

In this section, we present the core piece of our modeling approach, consisting of audio/image augmentation, pre-trained model development, tuning the pre-trained model, and discovering known/unknown patterns.

\subsubsection{Image vs. Audio Augmentations} 
As presented in Figure~\ref{model}, we can first compute the Log-Mel spectrogram images from audio clips and perform augmentations on images (i.e., {\em Image Augmentation} or IA approach). In this work, we consider two random image augmentation methods separately: (1) Random Cropping, where a random area from the image is cropped and resized to the original size, and (2) Gaussian Blur, which is a data smoothing method. On the other hand, we can first augment the audio events before computing the Log-Mel spectrograms (I.e., {\em Audio Augmentation} or AA approach). We use the open-source WavAug tool, which applies random pitch shift and room reverberation separately.

\ignore{-----------------------------
\subsubsection{Image Augmentation (IA)}\label{IA}
In this approach, audio events are first converted to Log-Mel spectrograms image features and then undergo two random image augmentation methods separately: (1) Random Cropping, where a random area from the log-Mel spectrogram is cropped and resized to the original size, and (2) Gaussian Blur, which is a data smoothing method.

\subsubsection{Audio Augmentation (AA)}\label{AA}
In this approach, the audio events are first augmented using the open-source WavAug tool, which applies random pitch shift and room reverberation separately. After this, the augmented audio events are transformed into Log-Mel spectrograms.
-----------------------------}

\subsubsection{Pre-trained Model Development}
The pre-trained model is used to extract a general-purpose representation from unlabelled data with a contrastive loss function. 
In the model shown in Figure~\ref{model}, ResNet-50 without the last fully connected layer is used as the encoder backbone to get representation vectors, e.g., $h_i$ and $h_j$ in the figure. Later, these vectors will serve as input for known and unknown pattern discovery models.
Using the SimCLR approach~\cite{chen2020simple}, the extracted vectors are passed through a small multi-layer perception (MLP) projection head with one hidden layer, which projects the vectors into an embedding space, e.g., $Z_i$ and $Z_j$ in the figure. Contrastive loss functions are applied to maximize the similarity between correlated instances. 

\ignore{-----------------------------
\subsubsection{Pre-trained Model}\label{pretrain}
The pre-trained model is used to extract a general-purpose representation from unlabelled data with a contrastive loss function. As is shown in Figure \ref{model} ,the augmented instances are fed into the neural network's base encoders, with ResNet-50 being chosen and the final FC layer removed as the backbone. Following the approach of SimCLR \cite{chen2020simple}, the extracted vectors are then passed through a small MLP network with one hidden layer that projects the representations into an embedding space. Here, contrastive loss functions are applied to maximize the similarity between correlated instances. After pre-training, the projection head is discarded, and the representations from the encoders are utilized for NDD tasks.
-----------------------------}

\subsubsection{Known and Unknown Pattern Discovery}
%
Next, we use the {\em open contrastive learning} (OpenCon) approach~\cite{sunopencon} to develop our unknown and known pattern discovery. The OpenCon model will identify four classes, i.e., Healthy, Flu, CC, and CB, where 90\% ``Healthy'' and ``Flu'' cough instances are labeled, and the remaining 10\% of the instances are unlabeled. Together they constitute the known patterns/classes. On the other hand, all CC and CB instances are unlabeled. 
The core idea behind OpenCon is to calculate a prototype vector for each known class and that vector will be updated during training. 
First, unlabeled instances from known and unknown classes will be compared with known labeled class prototypes using cosine similarities and then, the unlabeled instances will be distinguished into known or unknown classes based on the similarities. Next, the arguments of the maxima of cosine similarities between augmented embeddings and each prototype are calculated to get predicted labels. These augmented embeddings with the same predicted labels as anchor embedding are counted as positive sets to get new contrastive loss. After that, prototypes are updated. This is an iterative process.

%% file: analysis.tex
\section{Analysis}\label{analysis}

To analyze the performance of different pre-trained models developed with contrastive learning, we first train five random OpenCon models from each pre-trained model and test each of them on 10 random test sets of unlabeled unknown classes and labeled (90\%) and unlabeled (10\%) known classes. Train and test sets are always kept mutually exclusive based on users to avoid overfitting. In general, we observe higher accuracy when classifying unlabeled known class patterns compared to unlabeled unknown patterns, as expected. 
\textcolor{black}{Additionally, COVID-19 breathing (CB) is always better classified than COVID-19 coughing (CC), which can be due to the distinctiveness of breathing compared to known coughing classes. Similarly, ``Flu'' coughs are better identified than ``Healthy'' coughs, which may happen due to impurity in the publicly available Coswara dataset that contributes to both the unknown ``CC'' class and the known ``Healthy'' cough class.}
Next, we present our detailed analysis of different factors and their effects on models.

\ignore{-----------------------
Our study aims to investigate the potential effects of using different contrastive learning pre-trained models 
To this end, 
we trained each unknown pattern discovery model five times for each pre-trained model, and 
for each unknown pattern discovery model, we performed ten tests on different test subsets. Therefore, we analyzed fifty results for each pre-trained model.
-----------------------}

\begin{figure}[t]
\centering
\includegraphics[width=3.0in]{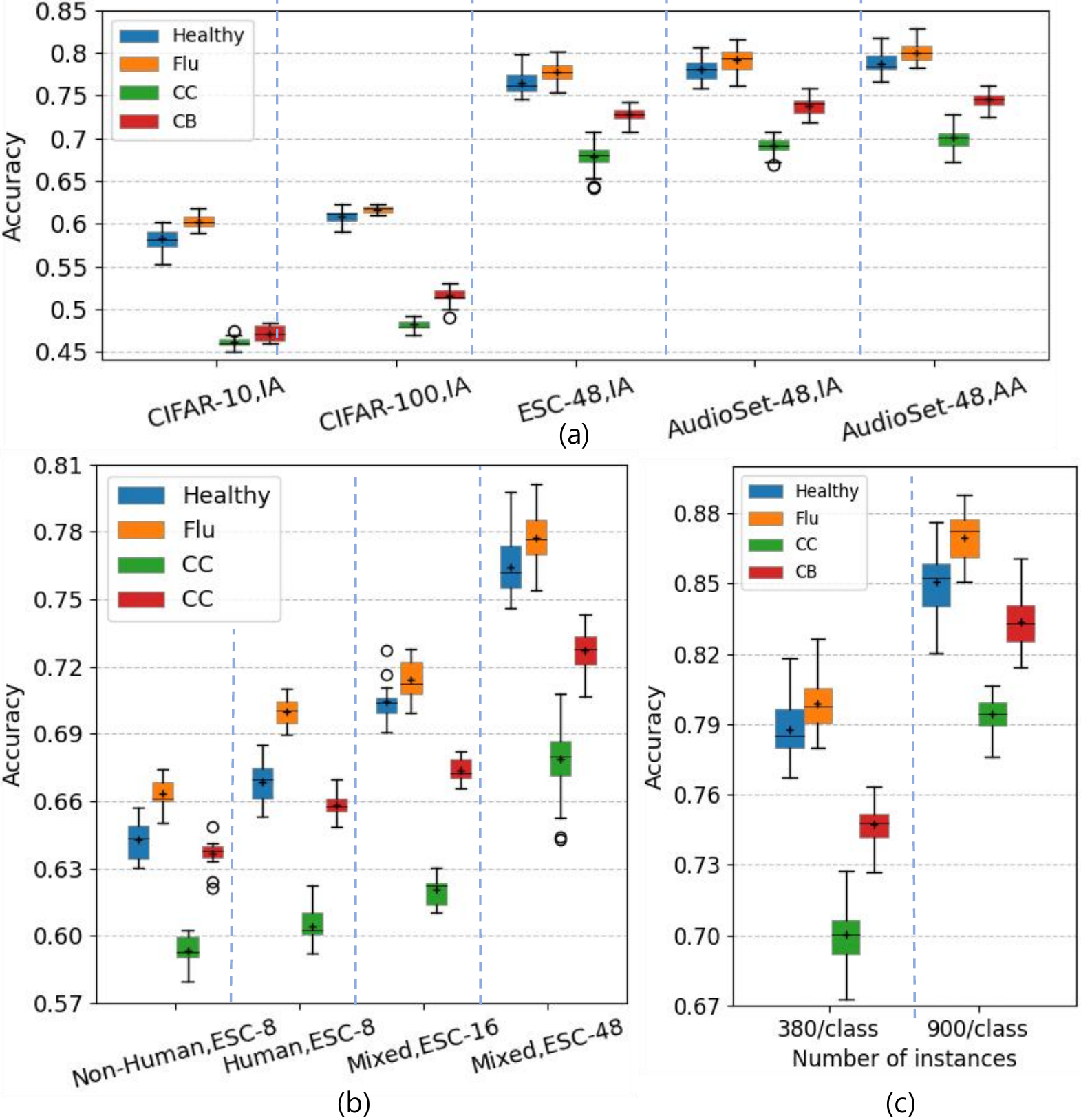}
\caption{Performance of pre-trained models developed from (a) different domains, (b) different levels of domain relevance (with IA), and (c) varying data volume of AudioSet-48 (with AA)}
\label{data}
\end{figure}

\subsection{Effect of Domains on the Pre-Trained Models}\label{domains}

In the left four sets of boxes in Figure~\ref{data}(a), we compare the performance of pre-trained models developed from the CIFAR dataset and two audio datasets, i.e., ESC and AudioSet using the IA approach with 380 instances per class. 
We observe that pre-trained models developed with audio domains, i.e., ESC or AudioSet, perform 19.7\% -- 44.6\% better than the models developed from image domains, i.e., object detection CIFAR dataset. Therefore, it is better to use data from domains similar to test domains when developing pre-trained models.   

Another interesting observation is AudioSet-48 data-driven pre-trained models outperform ESC-48 pre-trained models even with the same number of instances per class. This may link to the level of similarity among pre-trained classes and test classes since all classes in AudioSet-48 are human mouth or nose sounds compared to ESC-48 with 40 non-human sound classes. To investigate this further, we perform an additional experiment in Figure~\ref{data}(b). We develop pre-trained models with eight non-speech human sound classes (excluding coughing and breathing) and eight random non-human sound classes. Human data-driven ESC-8 pre-trained models perform 1.7\% - 5.7\% better than the Non-Human data-driven ESC-8 models due to relative closeness from the test known and unknown classes. 
Since the mixed ESC-16 (consisting of both Human ESC-8 and Non-Human ESC-8) has the best performance among the three with an average accuracy of $0.71\pm 0.01$ for known classes and $0.65\pm 0.01$ for unknown classes, we will further investigate the effect of data volume.

\subsection{Effect of Data Volume}\label{data volume}

On the right two sets of boxes in Figure~\ref{data}(b), we find that pre-trained models developed from the ESC-48 (i.e., 32 more non-human sound classes than ESC-16) outperform the ESC-16 pre-trained models by 17.6\% -- 21.5\% higher accuracy. This is probably because more classes provide more negative instances for contrastive learning, which contributes to producing a more generalized representation of instances.

Next, in Figure~\ref{data}(c), we investigate the effect of the number of instances per class on model performance by increasing the count from 380 to 900 for the AudioSet-48 dataset with the AA method. With around 2.4 times more instances per class, we witness 9.4\% -- 13.9\% higher accuracy, achieving an average accuracy of $0.86\pm 0.01$ for known classes and $0.81\pm 0.01$ for unknown classes. Therefore, an increase in the number of instances per class and the number of classes positively impact the model performance. 

\subsection{Effect of Augmentation Order}\label{order}

In the right two sets of boxes in Figure~\ref{data}(a), we investigate the effect of augmentation ordering, i.e., IA and AA approaches using the AudioSet-48 dataset with 380 instances per class. The AA approach is 1.1\% -- 2.4\%  more accurate than the IA approach when comparing the average accuracy values. 
This could be explained by the way these augmentation approaches work and their associated effect on augmented features. First, AA methods are better adapted to preserve the temporal characteristics of audio clips, i.e., AA methods can effectively maintain the temporal and sequential attributes of audio clips to learn valuable representations. On the other hand, some of the IA methods, such as random cropping and Gaussian blur, may eliminate crucial sequential information, given that image augmentation is aimed at capturing contour information with only certain important details. Second, AA methods may better reflect real-world variations in audio data compared to IA methods. For instance, audio augmentation techniques, such as pitch shifting, room reverberation, and random noise addition can more accurately simulate real-world scenarios, thereby enhancing the system's robustness when classifying audio recordings with varying qualities.

%% file: discussion.tex
\section{Discussion}\label{discussion}

Our study offers valuable insights into the design of a contrastive learning-based model for unknown pattern discovery tasks and key attributes of the pre-trained model that can affect performance. 
We find it is better to develop pre-training models with similar domains as the test domains and audio augmentations work better than image augmentations. 
Thereby, our work can contribute to the development of more effective and efficient methods to detect outbreaks of a new disease, which can have a significant impact on public health and the timely response to emerging infectious diseases.

Although our work has shown promising results, there are several limitations that we need to address in future research. First, the healthy cough instances in Coswara are from individuals who claim to have never been infected with COVID-19, but some of these individuals may have pre-existing respiratory conditions, such as asthma or long-term smoking habits that can possibly impact the model performance. Second, in our experiments, we treat COVID-19 coughing and breathing as two separate classes, but instances sometimes overlap, which adversely impacts model performances. Also, the unknown classes would probably be better to combine together since those symptoms are combined to diagnose COVID-19 in practice. 

Therefore, in our future work, we plan to conduct a large-scale longitudinal study with patients from widely varying demographic backgrounds and explore multi-modal fusion strategies for COVID-19 cough and breathing to optimize the model.